\documentclass[runningheads]{llncs}
\usepackage[T1]{fontenc}
\usepackage{graphicx}
\usepackage{hyperref}       
\usepackage{url}            

\usepackage{amsmath, amssymb, mathtools, amsfonts}
\usepackage{subcaption, array, booktabs}
\usepackage{color}

\usepackage[ruled]{algorithm2e}
\usepackage{algorithmic}
\usepackage{caption}
\usepackage{varioref}

\usepackage[capitalize,noabbrev]{cleveref}

\begin{document}
\title{Low-cost Robust Night-time Aerial Material Segmentation through Hyperspectral Data and Sparse Spatio-Temporal Learning}
%
%
\author{
Chandrajit Bajaj \inst{1,2}
\and Minh Nguyen \inst{1,2}
\and Shubham Bhardwaj \inst{1}
}

%
%
\institute{University of Texas at Austin
\and Equal contribution
}
%
\maketitle              
\begin{abstract}
Material segmentation is a complex task, particularly when dealing with aerial data in poor lighting and atmospheric conditions. To address this, hyperspectral data from specialized cameras can be very useful in addition to RGB images. However, due to hardware constraints, high spectral data often come with lower spatial resolution. Additionally, incorporating such data into a learning-based segmentation framework is challenging due to the numerous data channels involved. To overcome these difficulties, we propose an innovative Siamese framework that uses time series-based compression to effectively and scalably integrate the additional spectral data into the segmentation task. We demonstrate our model's effectiveness through competitive benchmarks on aerial datasets in various environmental conditions.

\end{abstract}
\section{Introduction}
Hyperspectral cameras capture the electromagnetic spectrum across multiple continuous bands, creating images with rich spectral information. This imaging technology has numerous important applications, including anomaly detection and remote sensing, where detailed spectral data can reveal insights that are not possible with conventional RGB imaging. For instance, Xu et al. \cite{anomaly_detection} use low-rank and sparse representations to decompose hyperspectral data into background and anomalous components. Similarly, Ojha et al. \cite{remotesensing} leverage hyperspectral data from the CRISM instrument to identify spectral signatures of hydrated salts on Mars, enabling them to detect and confirm the presence of these salts in recurring slope lineae.\\

\noindent However, hardware limitations often impose a trade-off between spatial and spectral resolutions, typically resulting in low spatial resolution for hyperspectral images (HSI). Despite this, applications like material segmentation can significantly benefit from low-resolution HSI. To address this challenge, we develop a novel framework that leverages low spatial resolution hyperspectral images (HSI) to improve material segmentation results. By employing innovative information compression techniques and advanced neural network architecture, we combine HSI data with noisy RGB image data to more accurately capture and classify materials at the pixel level. This framework enables the use of more affordable camera equipment while achieving performance comparable to high-resolution methods, which typically require expensive hyperspectral cameras.\\

Our specific contributions include:
\begin{enumerate}
    \item \textbf{Selective Channel Utilization}: We employ time series analysis to extract essential spectral data without processing the entire hyperspectral dataset, unlike low-rank methods that may lose crucial details or are limited to linear settings.
    
    \item \textbf{Advanced Deep Learning Architecture}: Our Siamese network integrates both hyperspectral and RGB data, in contrast to architectures that handle only a single data type.

    \item \textbf{Robustness to Adverse Conditions}: Our framework is built to perform reliably in challenging conditions, such as low lighting and atmospheric disturbances, whereas many existing methods target more standard scenarios.
\end{enumerate}

The rest of the paper is organized as follows: Section 2 provides an overview of material segmentation and the use of hyperspectral data for this task. Section 3 describes our detailed approach to handling the material segmentation task. Section 4 before the conclusion presents experimental results and analysis demonstrating the advantages of our approach.
 
\section{Background and related work}
\subsection{Material segmentation}
Material segmentation is a well-established task that extends image classification from the image level to the pixel level. \cref{fig:reconstruct} shows a labeling map for this task, featuring four different materials in four distinct colors within an aerial image \cite{jasper_ridge}. A significant amount of research has been conducted to address this problem in RGB images using various techniques: conditional random fields \cite{chen2014semantic}, convolutional network \cite{long2015fully}, structured prediction module \cite{zheng2015conditional,lin2016efficient,chen2017deeplab}, receptive field enlargement \cite{wu2018cgnet}, usage of boundary information \cite{li2020improving}, the contextual information refinement\cite{yu2020context}, attention modules \cite{xie2021segmenting}, and zero-shot generalization \cite{kirillov2023segment}.\\

\noindent While these methods focus exclusively on RGB images, there are few material segmentation works that extend beyond traditional RGB channels. Among these, \cite{salamati2010material} consider object-based segmentation with an additional channel near-infrared (NIR), while Liang et al.\cite{Liang_2022_CVPR} introduce the MCubeS dataset, that include only two additional channels, polarization and NIR. In their work, Liang et al. incorporate a region-guided filter selection (RGFS) layer to optimize the use of imaging modalities for each material class, resulting in effective classification. However, for aerial images such as JasperRidge \cite{jasper_ridge} or Urban \cite{jasper_ridge}, which contain hundreds of additional channels, there is a lack of research techniques to combine traditional RGB images with this extensive data.\\

\noindent New methods are needed to fully exploit the additional data collected from sophisticated imaging spectrometers with low spatial resolution. Recent advanced methods utilizing transformer architectures such as the work by Mazher et al.\cite{4d_cardio_imaging} represent a comprehensive approach that employs all available channels to capture both spatial and temporal features, ensuring no potentially useful data is discarded. While this approach exhaustively utilizes all data, our work differs by focusing on selective channel utilization to address specific challenges of hyperspectral imaging. Instead of processing all channels, our method identifies and retains only the most informative spectral channels, significantly reducing computational complexity while preserving critical information for accurate material segmentation.\\

\noindent From now on, we refer to the high-resolution images with only the three usual channels (red, green, blue) as RGB images, the hyperspectral images with low spatial resolution as HSI images, and the full hyperspectral images with high resolution as SRI images.

\subsection{Previous work on combining RGB and HSI images}
Several techniques have been proposed to combine RGB and HSI images to comprehensively understand the full SRI images underlying these pairs of RGB-HSI inputs. One class of methods \cite{pansharpening,bajaj2019blind} aims to enhance the spatial resolution of hyperspectral images through pansharpening techniques. Another approach, called hyperspectral unmixing, utilizes the underlying low-rank structure of the full SRI \cite{yokoya2017hyperspectral,lanaras2015hyperspectral}, as well as other prior information on the SRI's structure, such as spectral correlation \cite{CHEN2021102570}, sparsity information \cite{sparsityinfo}, or spatial degradation \cite{matrix_factorization,yokoya2012coupled,Dian2017}. \\

\noindent However, these methods are limited because they only handle individual images and are difficult to integrate into a deep learning framework for segmentation tasks. To address this problem, \cite{9706961} discusses a framework to train a model jointly on both HSI images and RGB images so that the auxiliary RGB super-resolution can provide additional supervision and regulate network training. Additionally, RGB images have been spectrally super-resolved with hyperspectral images using class-based backpropagation neural networks (CBPNNs) \cite{8615862}. Despite these advancements, these methods treat RGB and HSI images as separate entities, highlighting the need for a method that uses paired RGB-HSI inputs to enhance segmentation tasks. To this end, we develop a scalable technique that takes individual RGB-HSI pairs as inputs and outputs a desirable segmentation map.
\section{Proposed approach}
We formally describe the problem as follows:

\subsection{Problem description}
Our dataset $D$ consists of pairs of RGB and HSI images 
$D =\{(Z_i, Y_i)\}_{i = 1}^N$ with a total of $N$ image pairs. Here each RGB image $Z_i$ has dimension $3 \times H \times W$, where $3$ corresponding to the standard red, green, blue channels, and $H$ and $W$ are the (high resolution) height and width of the image. Similarly, each HSI image $Y_i$ has the dimension $C \times h \times w$ with (low resolution) height $h$ and width $w$, and the number of channels $C$. Here we assume that $H \gg h, W \gg w, C \gg 3$. We want to learn a material segmentation model defined by
\begin{equation}
\mathcal{S}: (Z_i, Y_i) \rightarrow M_i,
\end{equation} 
where $M_i$ is the predicted segmentation map for the $i^{th}$ input pair with the dimension $L \times H \times W$ such that $M_i(x, y)$ gives the predicted logits for the probabilities of occurrence (as an $L$-dimensional vector) over all material classes. We suppose that there are $L$ possible classes at each spatial pixel location $(x, y)$.

\subsection{General method}
\begin{figure}
    \centering
    {\includegraphics[scale=0.5]{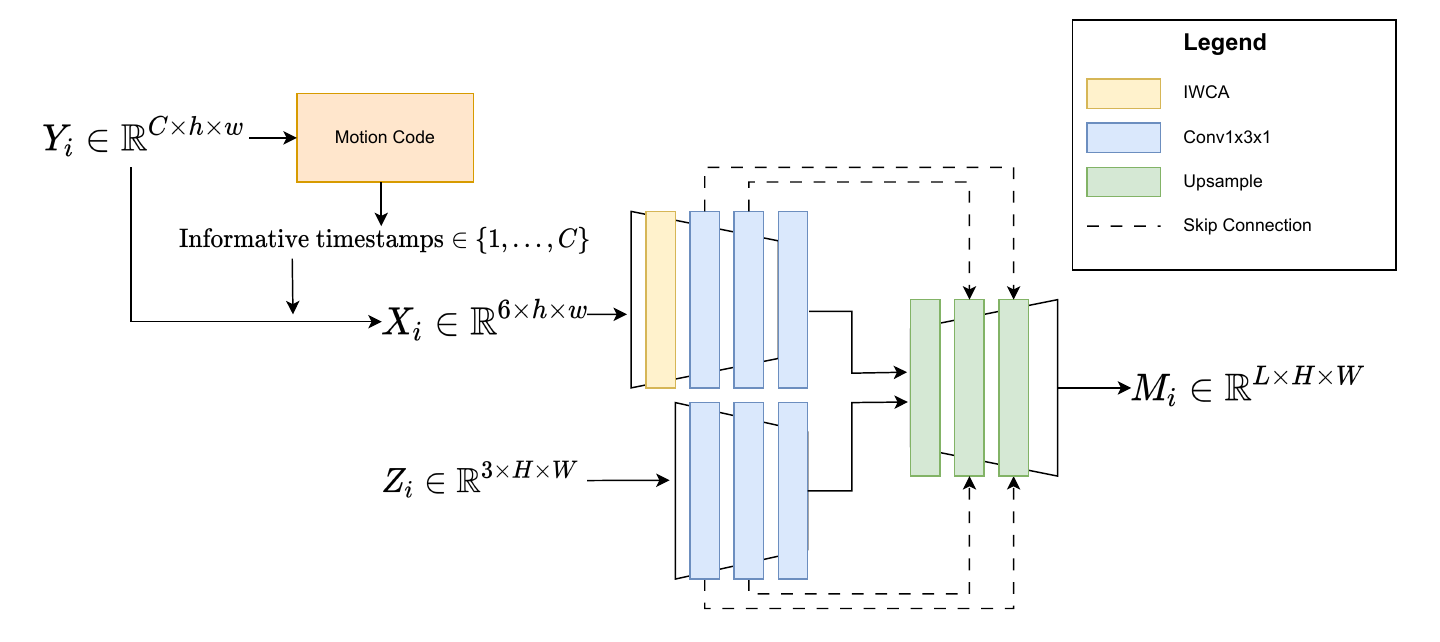}}
    \caption{General framework pipeline}\label{fig:architecture}
\end{figure}

Our framework (see \cref{fig:architecture}) includes two main steps:
\begin{enumerate}
    \item First we use the technique from \cite{motion_code} in order to extract the relevant information across $C$ channels to a smaller of size $6 \times h \times w$: $X_i = \mathcal{MC}(Y_i)$ (see \cref{sec:motion_code}), where $\mathcal{MC}$ is the function corresponding the extraction process.
    \item Then we feed the pair $(X_i, Z_i)$ into a Siamese network $\mathcal{S}$. This network is comprised of an IWCA module $I_{\phi}$ (see \cref{sec:iwca}), an encoder with 2 branches $E^{RGB}_{\theta}$ and $E^{HSI}_{\gamma}$, and a decoder $D_{\psi}$ (see \cref{sec:u_net} for encoder and decoder details). The final output has the form: 
    \begin{equation}\label{eq:main_net}
    \hat{M_i} = \mathcal{S}(X_i, Z_i) = D_{\psi}(E^{HSI}_{\gamma}(I_{\phi}(X_i)), E^{RGB}_{\theta}(Z_i))
    \end{equation}
\end{enumerate}

\begin{figure}[htb]
  \centering
  \subfloat[All series with labels]{\includegraphics[scale=0.3]{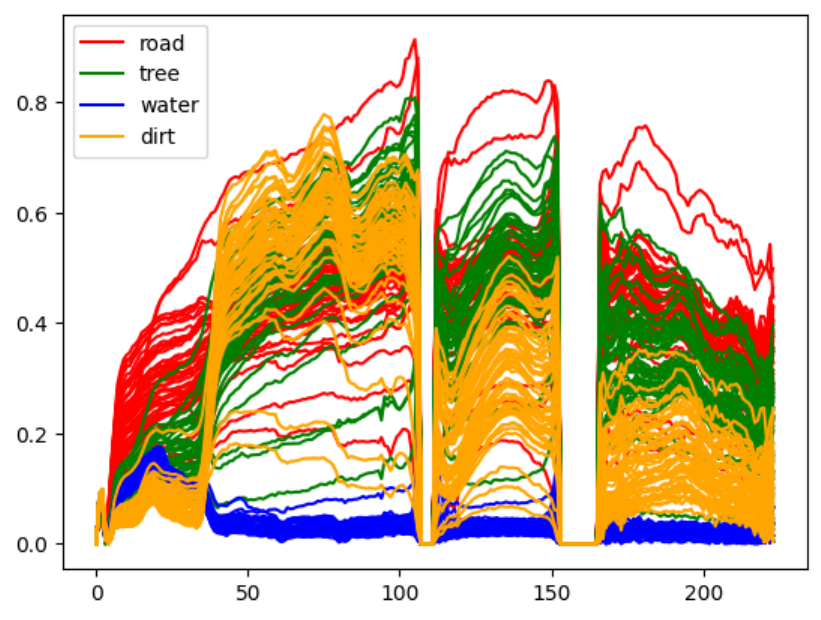}}
  \hfill
  \subfloat[\textbf{Road} materials]{\includegraphics[scale=0.3]{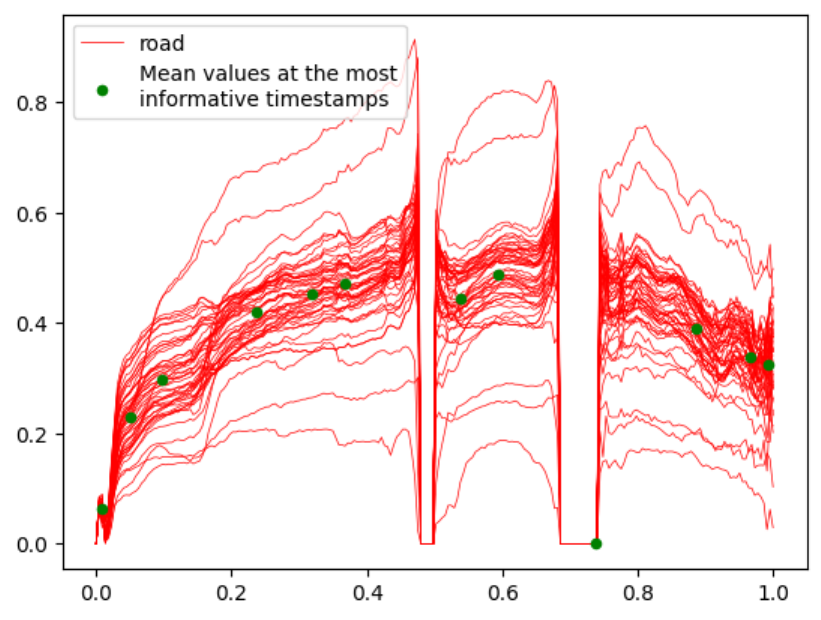}}
  \subfloat[\textbf{Dirt} materials]{\includegraphics[scale=0.3]{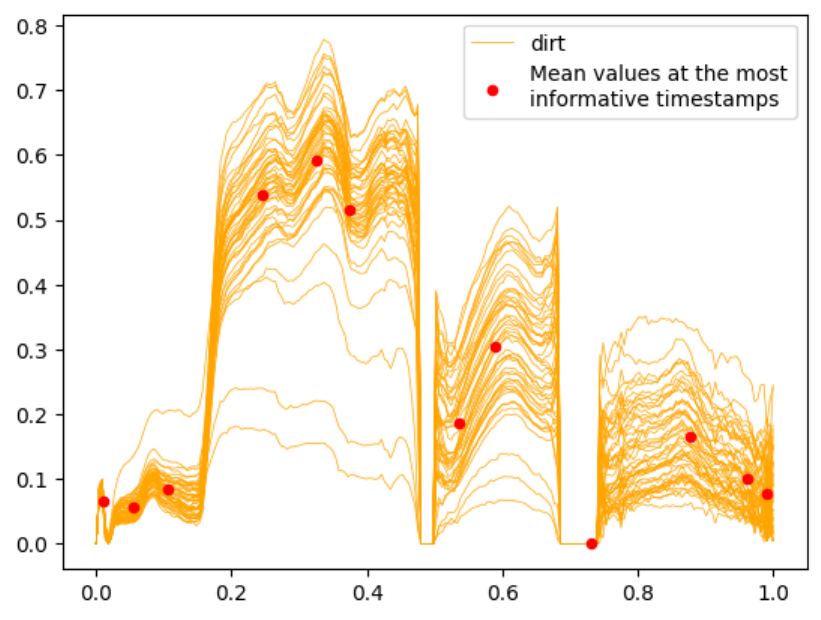}}
  \hfil
  \subfloat[\textbf{Water} materials]{\includegraphics[scale=0.3]{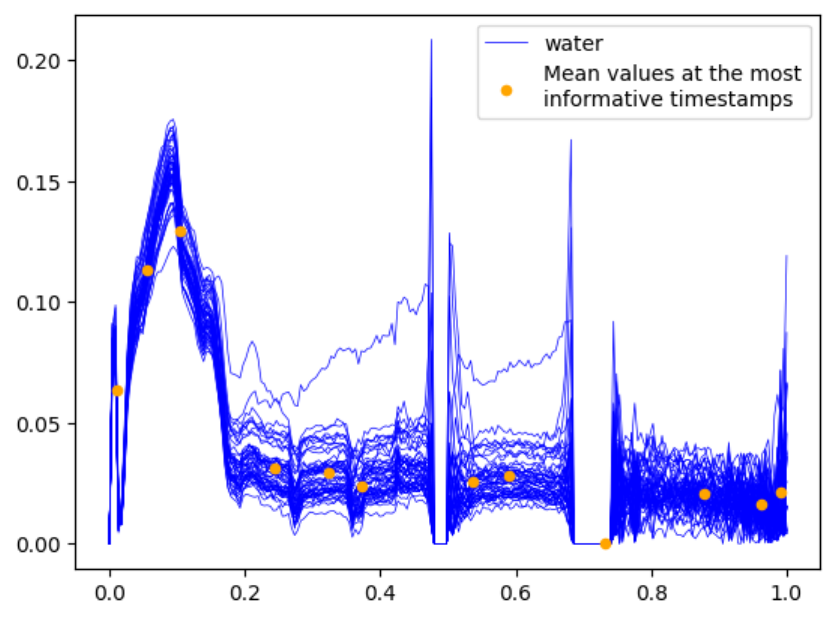}}
  \subfloat[\textbf{Tree} materials]{\includegraphics[scale=0.3]{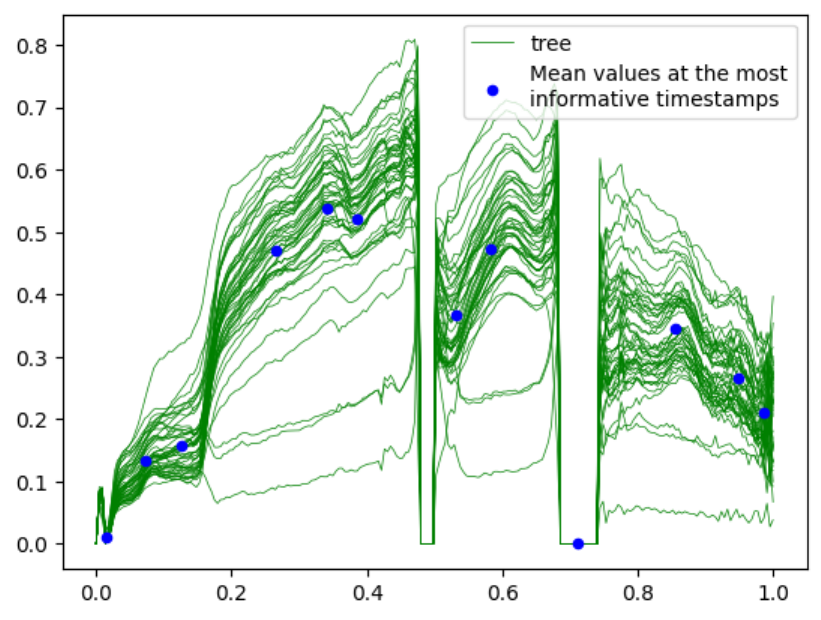}}
  \caption{The most informative timestamps and values for the timeseries induced from the original HSI image on each material class} \label{fig:most_informative_timestamp}
\end{figure}

\subsection{Selective learning for hyperspectral imaging}\label{sec:motion_code}
While HSI images are valuable for improving material prediction, their large number of channels makes deep learning training computationally intensive. To address this, our framework begins by reducing and condensing the relevant information for the segmentation task. Instead of learning from all $C$ channels, we select only the most informative channels for our hyperspectral input. Specifically, we first convert the original image data into time series data by considering pixel values across all channels. In other words, for each image in the training set and for each of the $h \times w$ pixel locations in that image, we create a time series of length $C$, where the values represent the corresponding pixel intensity. Each time series is then assigned a label based on the material class of its pixel (see \cref{fig:most_informative_timestamp}).\\

\noindent We then randomly select $n$ out of all possible time series. Next, we apply the time series analysis technique known as \textbf{Motion Code} from \cite{motion_code} to extract the most informative timestamps from this collection of time series across material classes. We refer to this entire process as the \textbf{fiber technique}. In our experiments, we selected $n = 200$. The informative timestamps are then used for band selection. Specifically, we choose the 12 most informative timestamps and refine them to retain only 6 channels by eliminating timestamps with near-zero intensity at both ends of the channel spectrum.\\

\begin{figure}[htb]
  \centering
  \subfloat[Original segmentation map]{\includegraphics[scale=0.5]{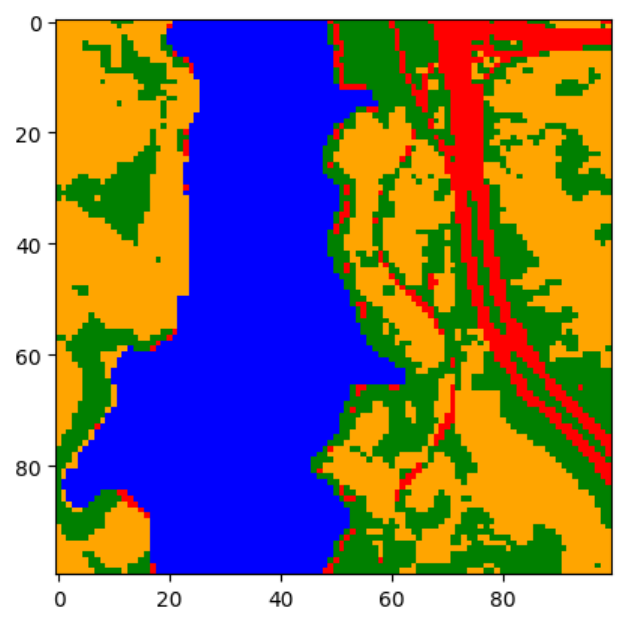}}
  \hfil
  \subfloat[Reconstructed segmentation map]{\includegraphics[scale=0.5]{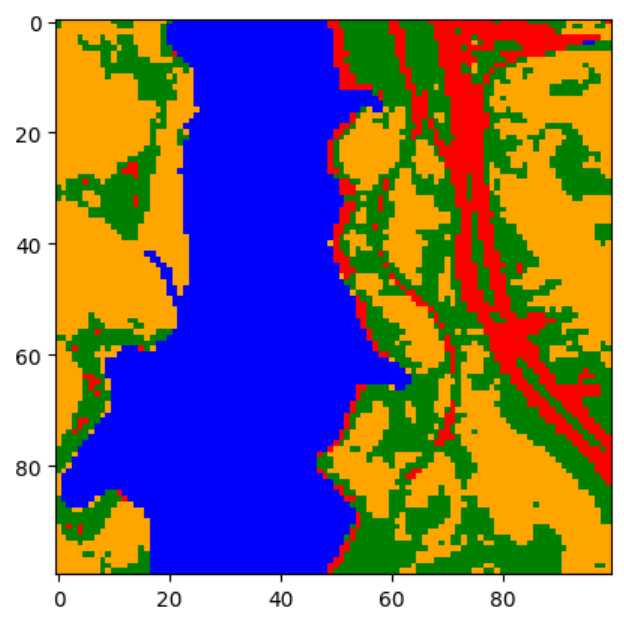}}
  \caption{Original and reconstructed material segmentation map using Motion Code} \label{fig:reconstruct}
\end{figure}

\noindent We reconstruct the material class segmentation using only these selected channels. To achieve this, we calculate the mean statistics over the most informative channels/timestamps from the small collection of $n = 200$ time series. Using these simple statistics alone, we can effectively reconstruct the global segmentation map shown in \cref{fig:reconstruct}. The near-perfect match between the predicted and original segmentation maps in \cref{fig:reconstruct} supports our claim that this small number of channels contains all the information necessary from the original $C$ channels for material segmentation.

\subsection{IWCA module} \label{sec:iwca}
Before entering the Siamese model, our extracted $X_i$ from the original HSI image $Y_i$ passes through a module called Importance Weighted Channel Attention (IWCA). Inspired by attention mechanisms \cite{attention}, the IWCA module re-weights the channels based on their importance and consists of two branches. The first branch uses regular $3 \times 3$ convolutions to integrate spatial and channel information. The second branch applies the group convolution operator to aggregate channel importance statistics, followed by average pooling and a sigmoid layer. Finally, the importance weight vector from the second branch is element-wise multiplied with the output of the first branch to produce the final output.

\subsection{Siamese U-net} \label{sec:u_net}
Our model's Siamese component leverages the strengths of both hyperspectral and RGB images to achieve high-quality segmentation results. The architecture is built on a modified U-Net structure, incorporating specialized blocks for downsampling, feature extraction, and upsampling.\\

\noindent The Siamese encoder is specifically designed to process both the low-resolution hyperspectral image (HSI) and the RGB image simultaneously. It features two parallel branches, each containing a series of convolutional layers that progressively downsample the input images and extract relevant features. For each input pair, the HSI image is processed by the IWCA module (see \cref{sec:iwca}) before entering one branch, while the RGB image is fed directly into the other branch. Each block in the encoder comprises a $1 \times 1$ convolutional unit, followed by a $3 \times 3$ convolutional unit, and another $1 \times 1$ unit. The $1 \times 1$ units handle channel mixing, while the $3 \times 3$ unit integrates both channel and spatial information.\\

\noindent The outputs from both branches are merged and then fed into the decoder, which consists of upsampling modules. These modules employ the bilinear upsampling to combine the feature maps from both the HSI and RGB encoder branches to produce the final segmentation map. The training loss function is defined as follows:
\begin{equation}
\mathcal{L}(\phi, \theta, \psi, \gamma) = \sum_{i=1}^{N_{train}}F^3(\hat{M_i}, T_i)
\end{equation}
Here $F^3$ is the focal loss function with fixed gamma parameter $3$ \cite{focal_loss}. The training set is $D_{train} = \{(Z_i, Y_i)\}_{i=1}^{N_{train}}$, and the segmentation prediction $\hat{M_i}$ is defined by \cref{eq:main_net} on sample RGB image $Z_i$ and HSI image $Y_i$. Lastly, $T_i$ is the truth material class label corresponding to the RGB image $Z_i$.

\section{Experiments}
\label{sec: Experiments}
\subsection{Datasets and experiment setup}
\begin{figure}[htb]
  \centering
  \subfloat[Original Image]{\includegraphics[scale=1]{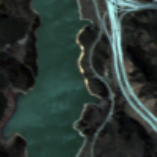}}
  \hfil
  \subfloat[Low-light]{\includegraphics[scale=1]{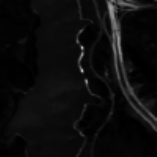}}
  \hfil
  \subfloat[Contrast\\Enhanced]{\includegraphics[scale=1]{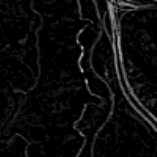}}
  \hfil
  \subfloat[Atmospheric\\Scattering]{\includegraphics[scale=1]{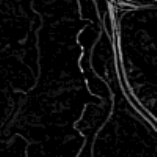}}
  \caption{Adverse image effects added incrementally from left to right on the \textbf{Jasper Ridge} dataset} \label{fig:aug_pipeline}
\end{figure}

We consider two aerial datasets \textbf{Jasper Ridge} and \textbf{Urban} \cite{jasper_ridge}. For each dataset, we extract pairs of HSI and RGB images through the following 4 steps:
\begin{enumerate}
    \item We first obtain $16 \times 16$ sub-views from the dataset's global scene by extracting $16 \times 16$ sub-images from the larger global scene. Each extracted sub-image represents an aerial view of a specific area within the captured regions. For example, in the Jasper Ridge dataset, one sub-image might capture an area near the river, while another might show a section deep in the jungle.
    \item For each of these sub-views, we obtain the corresponding RGB image with dimensions $3 \times 16 \times 16$. We also acquire the lower-resolution HSI image with dimensions $C \times 8 \times 8$ from the dataset's hyperspectral data.
    \item To account for the challenges encountered by hyperspectral equipment, we introduce low-light and atmospheric scattering effects (see \cref{adversity_effect}) to each sub-view image individually and independently. This approach simulates various real-life scenarios that hyperspectral devices might encounter (see \cref{fig:aug_pipeline} for an illustration).
    \item Applying this procedure to all sub-views results in the creation of the HSI-RGB dataset $D$, which consists of paired HSI and RGB images. Finally, we split this HSI-RGB dataset $D$ into disjoint training and test sets for model evaluation.
\end{enumerate}

We perform segmentation task on \textbf{Jasper Ridge} with 4 possible material classes: Road, Dirt, Water and Tree, and on \textbf{Urban} with 6 possible classes: Asphalt, Grass, Tree, Roof, Metal, and Dirt respectively. The number of channels is $C = 224$ for \textbf{Jasper Ridge}, and $C = 210$ for \textbf{Urban} dataset.

\subsection{Adversities effects}\label{adversity_effect}
\subsubsection{Low-Light effect Using Gamma Correction}
The low-light effect algorithm modifies a hyperspectral image by applying gamma correction to each spectral band. This involves normalizing the pixel values, applying the gamma transformation, and scaling them back to their original range. The process ensures that the image appears darker, simulating low-light conditions.

\subsubsection{Atmospheric scattering}
We adjust images based on a atmospheric scattering model that incorporates haze formation within images \cite{zhang2021density} with the following equation:
\begin{equation}
    I(x) = J(x)t(x) + A(1 - t(x))
\end{equation}
where $I(x)$ represents the haze image collected by the camera and $J(x)$ denotes the real reflected light on the surface of objects (normal image). Additionally, $t(x)$ is the medium transmittance, $A$ is the atmospheric light intensity, and $x$ represents the pixel coordinate. Given the image depth map $d(x)$ and the atmospheric scattering coefficient $\beta$, the transmittance $t(x)$ can be calculated as $t(x) = e^{-\beta d(x)}$. We use random noise to simulate depth, and set $\beta = 0.1$, $A = 0.8$ in our experiments.

\subsection{Evaluation results}
We select two other baselines: U-Net \cite{U_net} and convolutional neural network (CNN) models trained solely on RGB images to assess the impact of our approach in incorporating low-resolution hyperspectral data. We have 2 evaluation metrics:
\begin{enumerate}
    \item Per-class mean IOU (mIOU), where for each material class, we compute the mean of all IOU \cite{Liang_2022_CVPR} over all pixels.
    \item The generalized dice score (gDice) metric across all material classes. \cite{sudre2017generalised}
\end{enumerate}

\begin{table}[ht]
\caption{Material segmentation results on \textbf{Jasper Ridge} dataset, with only atmospheric scattering effects.} 
\centering
\vspace{2mm}
\begin{tabular}{|c|c|c|c|c|c|}
    \hline
    \text{Model} & \multicolumn{4}{|c|}{Classwise mIOU} & \text{gDice} \\
    \cline{2-5}
    & \parbox{32pt}{\centering Road} & \parbox{32pt}{\centering Dirt} & \parbox{32pt}{\centering Water} & \parbox{32pt}{\centering Tree} &  \\
    \hline
    U-Net & 0.4966 & 0.6429 & \color{red}{0.5889} & 0.7069 & 0.8920 \\
    CNN & 0.4320 & 0.5886 & 0.5860 & 0.6733 & 0.8546 \\
    Ours & \color{red}{0.5012} & \color{red}{0.6597} & 0.5886  & \color{red}{0.7159} & \color{red}{0.9022} \\
    \hline
\end{tabular}
\label{tab:jasper_ridge}
\end{table}

Our method outperforms other baseline models, as demonstrated in \cref{tab:jasper_ridge}, when only the atmospheric scattering effect (flight effect on day time) is applied. More importantly, we present results showcasing our method's performance under low-light night-time scenarios, with and without contrast enhancement. Specifically, our method significantly surpasses the other baselines, not only in terms of gDice but also in mIOU for each class, across both the Jasper Ridge and Urban datasets (see \cref{tab:jasper_ridge2} and \cref{tab:urban_dataset}). 

\begin{table}[ht]
\caption{Material segmentation results on \textbf{Jasper Ridge} dataset with both atmospheric scattering and contrast enhancement effects for night time flight scenarios}
\centering
\vspace{2mm}
\begin{tabular}{|c|c|c|c|c|c|c|}
    \hline
    \text{Model} & \parbox{50pt}{\centering Contrast Enhance} & \multicolumn{4}{|c|}{Classwise mIOU} & \text{gDice} \\
    \cline{3-6}
    & & \parbox{32pt}{\centering Road} & \parbox{32pt}{\centering Dirt} & \parbox{32pt}{\centering Water} & \parbox{32pt}{\centering Tree} &  \\
    \hline
    Ours & Y & \color{red}{0.3454} & \color{red}{0.5228} & \color{red}{0.5652} & \color{red}{0.6239} & \color{red}{0.7887} \\
    Ours & N & 0.2685 & \color{blue}{0.4850} & \color{blue}{0.5586} & \color{blue}{0.6001} & \color{blue}{0.7609} \\
    U-Net & Y & \color{blue}{0.2929} & 0.4396 & 0.5398 & 0.5604 & 0.6415 \\
    U-Net & N & 0.2235 & 0.2855 & 0.4658 & 0.4097 & 0.3706 \\
    CNN & Y & 0.1587 & 0.3892 & 0.5412 & 0.5158 & 0.6737 \\
    CNN & N & 0.1411 & 0.2894 & 0.4785 & 0.3998 & 0.4992 \\
    \hline 
\end{tabular}
\label{tab:jasper_ridge2}
\end{table}

\subsubsection{Contrast Enhancement}
Images taken in low-light conditions often suffer from poor contrast, making it difficult to extract meaningful information. In addition to the regular setting, we also allow contrast enhancement through a series of processing steps \cite{yu2004fast}. The preprocessing includes Gaussian filtering for anisotropic propagation controlled by the standard deviation $\sigma$, also known as the conductivity coefficient $C$, and computing local statistics (minimum, maximum, and average) within a sliding window. We also apply a parabolic transfer function to each pixel, adjusting the contrast based on its intensity relative to the local minimum, maximum, and average.

\begin{table}[ht]
\caption{Segmentation results on Urban dataset with Atmospheric scattering and low-light effect to simulate night time flight scenario}
\centering
\vspace{2mm} 
\begin{tabular}{|c|c|c|c|c|c|c|c|c|}
    \hline
    \text{Model} & \parbox{50pt}{\centering Contrast Enhance} & \multicolumn{6}{|c|}{classwise mIOU} & \text{gDice} \\
    \cline{3-8}
    & & \parbox{32pt}{\centering Asphalt} & \parbox{32pt}{\centering Grass} & \parbox{32pt}{\centering Tree} & \parbox{32pt}{\centering Roof} & \parbox{32pt}{\centering Metal} & \parbox{32pt}{\centering Dirt} &  \\
    \hline
    Ours & Y & \color{red}{0.7775} & \color{red}{0.7982} & \color{red}{0.7556} & \color{red}{0.5972} & \color{red}{0.4461} & \color{red}{0.6094} & \color{red}{0.6150} \\
    Ours & N & \color{blue}{0.7421} & \color{blue}{0.7894} & \color{blue}{0.7266} & \color{blue}{0.5382} & \color{blue}{0.3630} & \color{blue}{0.5775} & \color{blue}{0.5451} \\
    U-Net & Y & 0.7004 & 0.6981 & 0.5321 & 0.4932 & 0.3388 & 0.5131 & 0.4882 \\
    U-Net & N & 0.7006 & 0.7297 & 0.6465 & 0.4818 & 0.3281 & 0.5219 & 0.4903 \\
    CNN & Y & 0.6409 & 0.6583 & 0.4547 & 0.3952 & 0.2913 & 0.4023 & 0.4263 \\
    CNN & N & 0.5638 & 0.6168 & 0.4741 & 0.3448 & 0.2244 & 0.3385 & 0.3591 \\
    \hline 
\end{tabular}
\label{tab:urban_dataset}
\end{table}

\subsection{Night-time effect analysis} \label{sec:night_time}

\begin{figure}[!ht]
    \centering
    \includegraphics[scale=0.4]{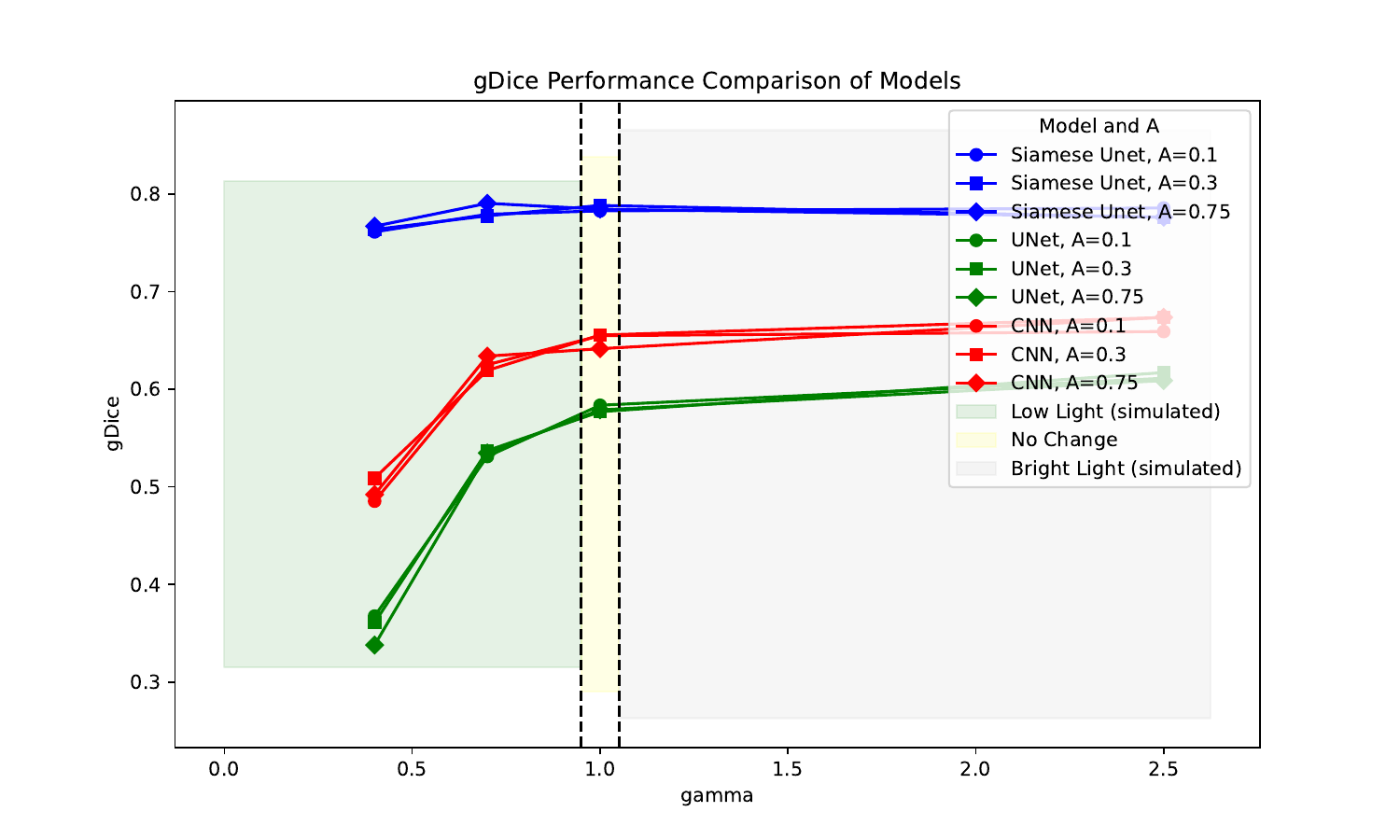}
    \caption{\textbf{gDice} performance comparison of models with varying levels of darkness and atmospheric scattering on \textbf{Jasper Ridge} dataset, higher $A$ leads to higher scattering. Our model (blue) performs consistently. Meanwhile, degradation of RGB only models in low-light scenarios is drastic.}
    \label{fig:gdice_performance}
\end{figure}

\begin{figure}[!ht]
    \centering
    \includegraphics[scale=0.4]{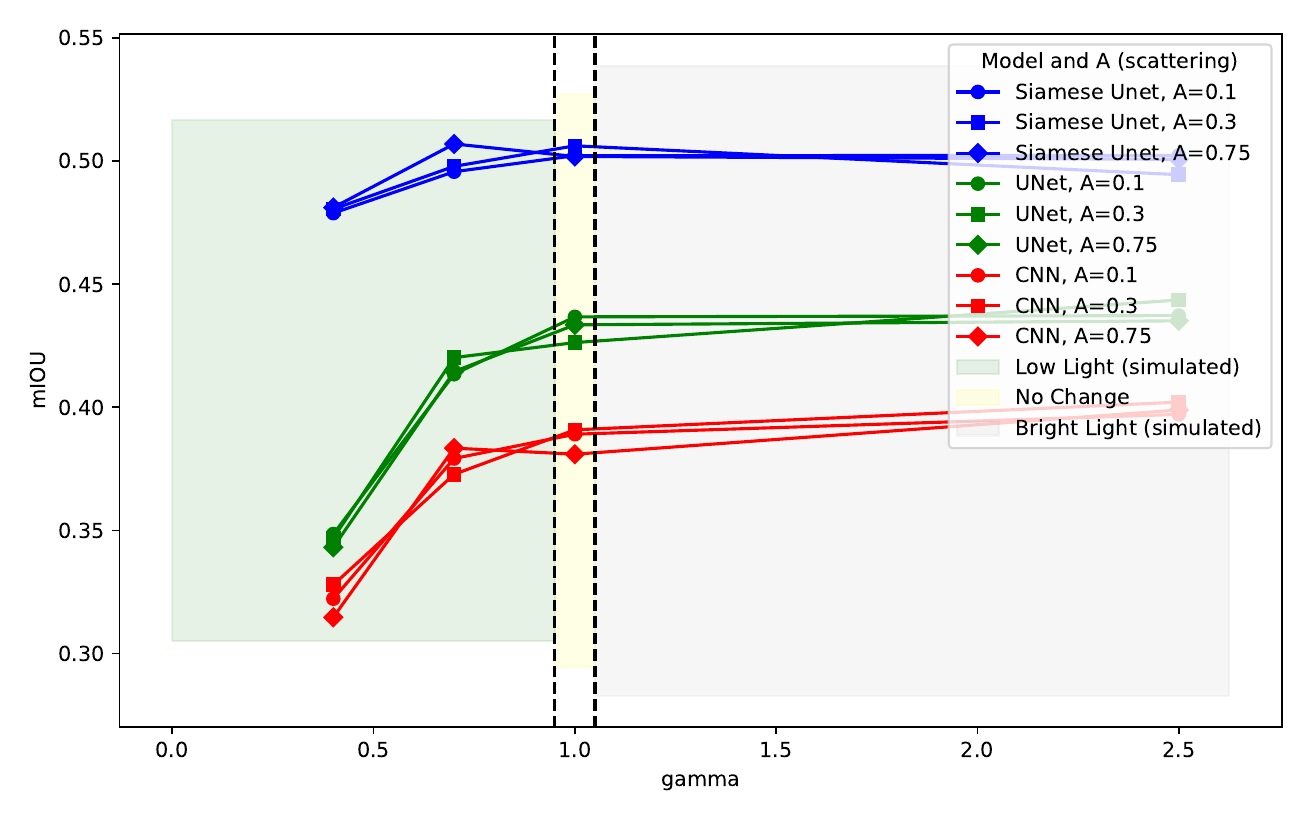}
    \caption{\textbf{mIOU} performance comparison of models with varying levels of darkness and atmospheric scattering on \textbf{Jasper Ridge}, higher $A$ leads to higher scattering. Our model (blue) performs consistently. Meanwhile, degradation of RGB only models in low-light scenarios is drastic}
    \label{fig:miou_performance}
\end{figure}

\noindent Our model significantly outperforms conventional RGB-based models like CNN and U-Net in low-light conditions, as shown in the \cref{fig:gdice_performance} and \cref{fig:miou_performance}. RGB-only models struggle under night-time scenarios due to their reliance on visible light, which is severely limited in such conditions. As a result, these models show poor segmentation accuracy, particularly in distinguishing material boundaries like roads or water surfaces.\\

\noindent In contrast, our Siamese network consistently maintains high accuracy by incorporating hyperspectral data alongside RGB inputs. Hyperspectral channels, especially those in the infrared (IR) spectrum, capture crucial material properties that remain visible even in low-light environments. These channels allow the model to effectively differentiate between materials based on their reflectance and surface textures, which are otherwise indistinguishable in darkness using RGB data alone. This capability gives our model a significant advantage over traditional approaches, allowing it to ``see through'' low-light conditions and provide more accurate material segmentation.\\

\noindent Finally, the model's robustness to adverse conditions, such as atmospheric scattering, is further enhanced by its selective channel usage to focus on relevant information while minimizing computational complexity. Such approach enables the model to perform consistently well, even under varying levels of darkness and scattering, as seen in the performance comparison.

\subsection{Visualization analysis}\label{sec:visual}
\noindent Visualization of the segmentation maps generated by our model versus the baseline models clearly highlights the strengths of our approach (see \cref{tab:zoomed}). In challenging areas, both CNN and U-Net models struggle to maintain the topological structure of the scene. In contrast, our model preserves the geometric integrity of the materials being segmented. \\

\noindent This is especially evident in regions with mixed materials, such as transitions between road and dirt or water and tree areas. The U-Net and CNN models introduce substantial noise and errors in these regions, resulting in significant misclassifications, as seen in the highlighted areas in \cref{tab:zoomed}. The visualizations highlight our model’s effectiveness in maintaining structural coherence in the segmentation maps. This capability is critical for applications where understanding the topology and material distribution in aerial imagery is essential.

\section{Conclusion}
In this paper, we developed a framework that combines traditional RGB images with hyperspectral data to tackle the material segmentation task on aerial data, particularly under night-time and challenging atmospheric conditions. We demonstrated our framework's capabilities and robustness in improving segmentation results through the efficient use of hyperspectral data. In the future, we aim to extend our work to more general datasets and other critical tasks beyond material segmentation.\\

The implementation is available at \href{https://github.com/CVC-Lab/HSI-MSI-Image-Fusion}{https://github.com/CVC-Lab/HSI-MSI-Image-Fusion}.

\begin{credits}
\subsubsection{\ackname} This research was supported in part from the Peter O’Donnell Foundation, the Michael J Fox Foundation, and Jim Holland-Backcountry Foundation.

\noindent This preprint has no post-submission improvements or corrections. The Version of Record of this contribution is published in the Neural Information Processing, ICONIP 2024 Proceedings.
\end{credits}

\begin{table}[htbp]
    \centering
    \captionsetup{width=\textwidth}
    \begin{tabular}{cccc}
        \textbf{Ground Truth} & \textbf{CNN} & \textbf{U-Net} & \textbf{Ours} \\[10pt]
        \begin{minipage}[b]{0.25\linewidth}
            \centering
            \includegraphics[width=\linewidth]{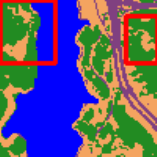}
        \end{minipage} &
        \begin{minipage}[b]{0.25\linewidth}
            \centering
            \includegraphics[width=\linewidth]{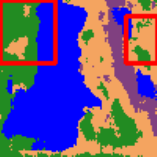}     
        \end{minipage} &
        \begin{minipage}[b]{0.25\linewidth}
            \centering
            \includegraphics[width=\linewidth]{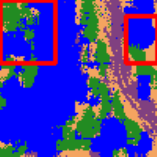}
        \end{minipage} &
        \begin{minipage}[b]{0.25\linewidth}
            \centering
            \includegraphics[width=\linewidth]{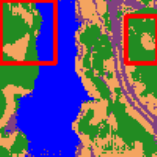}       
        \end{minipage} \\[10pt]
         \begin{minipage}[b]{0.25\linewidth}
            \centering
            \includegraphics[width=\linewidth]{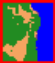}
        \end{minipage}
         &
        \begin{minipage}[b]{0.25\linewidth}
            \centering
            \includegraphics[width=\linewidth]{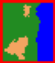} 
        \end{minipage} &
        \begin{minipage}[b]{0.25\linewidth}
            \centering
            \includegraphics[width=\linewidth]{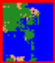}
        \end{minipage} &
        \begin{minipage}[b]{0.25\linewidth}
            \centering
            \includegraphics[width=\linewidth]{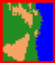}    
        \end{minipage}\\[10pt]
         \begin{minipage}[b]{0.25\linewidth}
            \centering
            \includegraphics[width=\linewidth]{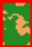}
        \end{minipage}
         &
        \begin{minipage}[b]{0.25\linewidth}
            \centering
            \includegraphics[width=\linewidth]{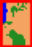}
        \end{minipage} &
        \begin{minipage}[b]{0.25\linewidth}
            \centering
            \includegraphics[width=\linewidth]{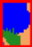}
        \end{minipage} &
        \begin{minipage}[b]{0.25\linewidth}
            \centering
            \includegraphics[width=\linewidth]{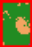}    
        \end{minipage}
    \end{tabular}
    \caption{\textbf{First row}: Ground truth and predicted material class labels from our models and other baselines. Two red regions are the regions to be zoomed in for further inspection in second and third rows. \textbf{Color}: Road: Purple, Dirt/Soil: Brown, Water: Blue, Tree: Green.\\
    \textbf{Second row}: Left bounding box segmentation map zoomed in.\\
    \textbf{Third row}: Right bounding box segmentation map zoomed in}
    \label{tab:zoomed}
\end{table}

%
%
%
\bibliographystyle{splncs04}
\bibliography{citation}
%




\end{document}